\documentclass{article}

\usepackage{microtype}
\usepackage{graphicx}
\usepackage{subcaption}
\usepackage{booktabs} 

\usepackage{hyperref}

\usepackage[accepted]{icml2026}

\usepackage{amsmath}
\usepackage{amssymb}
\usepackage{mathtools}
\usepackage{amsthm}

\usepackage[capitalize,noabbrev]{cleveref}

\theoremstyle{plain}

\theoremstyle{definition}

\theoremstyle{remark}

\usepackage[textsize=tiny]{todonotes}

\icmltitlerunning{Multi-task Learning is Not Enough}

\begin{document}

\twocolumn[
  \icmltitle{Multi-task Learning is Not Enough: Representational Entanglement \\ in Dual-output Second Language Speech Recognition}




  \begin{icmlauthorlist}
    \icmlauthor{Seung Hwan Cho}{yyy1}
    \icmlauthor{Young-Min Kim}{yyy1,yyy2}
  \end{icmlauthorlist}

  \icmlaffiliation{yyy1}{Department of Industrial Data Engineering, Hanyang University, Seoul, South Korea}
  \icmlaffiliation{yyy2}{School of Interdisciplinary Industrial Studies, Hanyang University, Seoul, South Korea}

  \icmlcorrespondingauthor{Young-Min Kim}{yngmnkim@hanyang.ac.kr}

  \icmlkeywords{Machine Learning, ICML}

  \vskip 0.3in
]



\printAffiliationsAndNotice{}  

\begin{abstract}
Second-language (L2) speech recognition often requires  transcriptions of pronunciations and intended meanings.  
Multi-task learning (MTL) is a natural approach because it  assumes that shared representations benefit both outputs.  
However, this paper shows that this assumption does not hold across Korean and English. 
MTL improves meaning but degrades surface transcription, especially in English, where the degradation scales with surface-meaning divergence measured by Levenshtein edit distance. 
Encoder analysis links these patterns to encoder-level entanglement, with Korean preserving disentangled representations while English produces nearly  identical ones.
Cross-output decoder analysis shows that the meaning dual-output decoder adapts with a unique representation, while the surface dual-output decoder remains constrained by the encoder.
These findings motivate the design of MTL frameworks that mitigate encoder-level entanglement to reduce surface degradation in dual-output L2 automatic speech recognition.
\end{abstract}

\section{Introduction and Related Work}

Human speech exhibits systematic differences between what is actually pronounced (surface-level) and the canonical written form (meaning-oriented) of an utterance.
These differences reflect language-specific phonological phenomena, such as coarticulation, phonological reduction, and liaison \citep{ernestus2011introduction}. 
The phonological gap is more pronounced in second-language (L2) speech, where speaker-specific deviations are common \citep{munro2021difficulty}. 
Therefore, speech recognition systems for L2 speakers must recover both transcription forms from a single acoustic signal to enable targeted feedback in language learning and pronunciation assessment applications \citep{eskenazi2009overview}.

Multi-task learning (MTL) is a natural approach for dual-output (DO), encompassing auxiliary MTL where one task supports another and joint MTL where tasks are learned with equal importance \citep{ruder2017overview}. 
In automatic speech recognition (ASR), joint connectionist temporal classification (CTC)-attention training~\citep{kim2017joint,watanabe2017hybrid} and intermediate-layer CTC~\citep{nozaki2021relaxing} use auxiliary CTC to improve alignment and training stability.
Joint MTL approaches generate distinct target sequences from the same acoustic input. 
Examples include dual-decoder models for ASR and speech translation \citep{le2020dual} and unified diarization-separation-recognition systems \citep{shakeel2025unifying}.
However, the effectiveness of joint MTL for DO L2 ASR remains unexamined, despite the fact that the two outputs share linguistic content.

This paper challenges the assumption that joint MTL benefits both outputs in DO ASR by conducting controlled experiments on Korean and English L2 speech. 
The results show that joint MTL produces asymmetric output trade-offs that depend on language, with the source localized to encoder-level representational entanglement. 
Our contribution is twofold. 
First, we demonstrate the language-dependent behavior of joint MTL for DO L2 ASR. 
Second, we identify the underlying mechanisms through encoder and cross-output decoder analyses.
These findings motivate the development of structured approaches to mitigate this entanglement.

\begin{figure*}[t]
\centering
\includegraphics[width=0.9\textwidth]{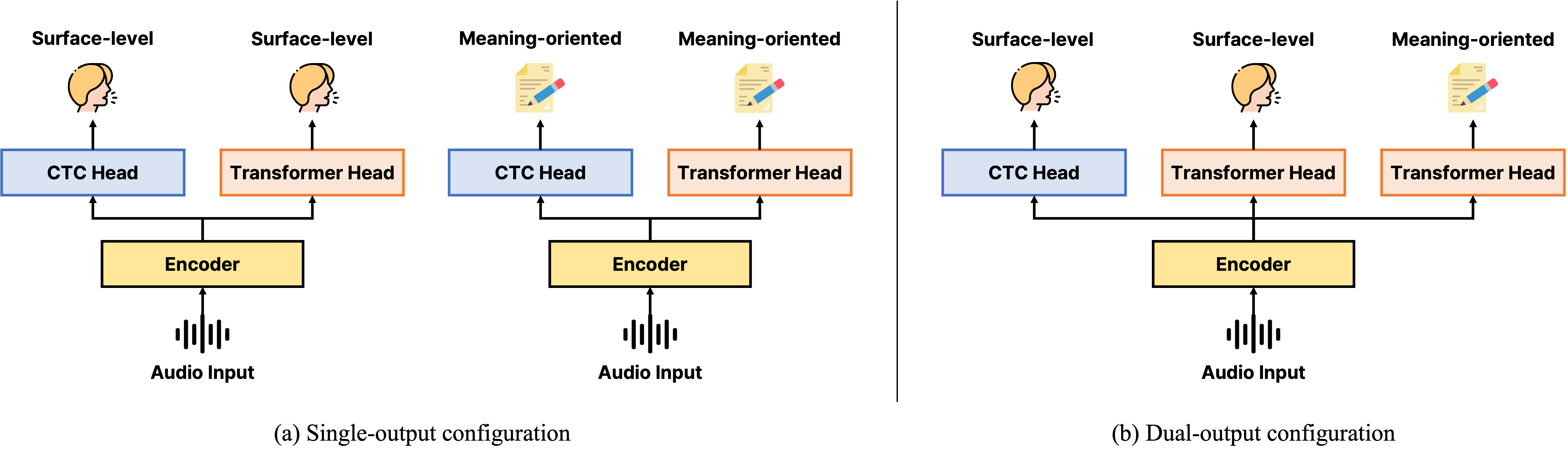}
\caption{(a) Single-output configuration trains separate models for surface-level (left) and meaning-oriented (right) transcription, each with its own encoder, decoder, and auxiliary CTC head. 
(b) Dual-output configuration shares a single encoder with two separate Transformer decoders that jointly produce both outputs, along with an auxiliary CTC head on the encoder output.}
\label{fig:architecture}
\end{figure*}

\section{Method}
To isolate the effect of joint training on shared representations, we compare single-output (SO) models, which are trained separately on each output, and DO models, which are trained jointly. 
This comparison is illustrated in Figure~\ref{fig:architecture}.

\subsection{Problem Formulation}
Given an input log mel-spectrogram $X \in \mathbb{R}^{T \times 80}$ with $T$ frames, two output sequences are produced.
The surface-level transcription $y^{\text{surf}}$ represents the verbatim spoken form, while the meaning-oriented transcription $y^{\text{mean}}$ represents the intended written form. 
Each token is drawn from a fixed vocabulary $\mathcal{V}$.

\subsection{Architecture and Training}
\paragraph{Single-output Model.}
Figure~\ref{fig:architecture}(a) shows the SO baseline, which follows the hybrid CTC-attention paradigm with an auxiliary CTC head on the encoder output for alignment supervision. 
Separate models are trained for surface-level and meaning-oriented transcription. 
The training objective is $\mathcal{L}_{\text{single}} = \alpha \mathcal{L}_{\text{CTC}} + (1-\alpha) \mathcal{L}_{\text{att}}$ with $\alpha$ = 0.2, where $\mathcal{L}_{\text{CTC}}$ is the CTC loss on the target transcription, $\mathcal{L}_{\text{att}}$ is the cross-entropy loss of the attention decoder.

\paragraph{Dual-output Model.}
The DO model uses one encoder and two decoders, one for each transcription type shown in Figure~\ref{fig:architecture}(b). 
Both decoders attend to the same encoder output through audio cross-attention, and the auxiliary CTC head is retained on the encoder output. 
The auxiliary CTC is trained on surface-level targets, which preserve the monotonic frame-token alignment that CTC assumes. 
The training objective is $\mathcal{L}_{\text{dual}} = \alpha \mathcal{L}_{\text{CTC}} + \beta \mathcal{L}_{\text{surf}} + \gamma \mathcal{L}_{\text{mean}}$ with $(\alpha, \beta, \gamma) = (0.2, 0.5, 0.3)$, all fixed via pre-experiments on the validation set. 
By keeping the auxiliary CTC identical in both configurations, the only architectural difference lies in the presence of the second decoder.

\begin{table}[h]
\centering
\small
\caption{Dataset statistics and Distribution of surface-meaning divergence using Levenshtein edit distance.}
\label{tab:dataset}
\setlength{\tabcolsep}{12pt} 
\begin{tabular}{lrr}
\toprule
& \textbf{Korean} & \textbf{English} \\
\midrule
Train       & 33,442 & 57,616 \\
Validation  &  4,180 &  7,199 \\
Test        &  4,181 &  7,207 \\
\textbf{Total} & \textbf{41,803} & \textbf{72,022} \\
\midrule
\multicolumn{3}{l}{\textit{Surface--meaning edit distance}} \\
ED $=$ 0          & 16,203 (38.8\%) & 18,774 (26.1\%) \\
ED $=$ 1--3       & 14,790 (35.4\%) & 28,993 (40.3\%) \\
ED $=$ 4--10      &  8,617 (20.6\%) & 21,256 (29.5\%) \\
ED $\geq$ 11      &  2,193 (5.2\%)  &  2,999 (4.2\%)  \\
\bottomrule
\end{tabular}
\end{table}

\section{Experiments}
\subsection{Experimental Setup}
\paragraph{Dataset.}
The two AI-Hub\footnote{This research used datasets from `The Open AI Dataset Project (AI-Hub, S. Korea)'. All data information can be accessed through AI-Hub (\url{www.aihub.or.kr}).} datasets, "Educational Korean Audio Data Recorded by Native (L1) Chinese and Japanese Speakers" and "Educational English Audio Data Recorded by L1 Korean Speakers," which include all read speech samples, are used.
Table~\ref{tab:dataset} summarizes the dataset statistics and surface-meaning divergence distribution. 
The Korean dataset contains 41,803 samples, while the English dataset contains 72,022 samples. 
Surface-meaning divergence is measured by the Levenshtein edit distance (ED) between surface and meaning transcriptions, with character-level syllables for Korean and word-level tokens for English. 
Both languages exhibit similar divergence distributions, with the majority of samples falling in the ED 0--3 range and only a small portion exceeding ED$=$10.

\paragraph{Baselines and Implementation Details.}
For SO, Whisper base and small~\citep{radford2022robustspeechrecognitionlargescale} fine-tuned from pretrained weights and Conformer \citep{gulati2020conformer} with a single transformer decoder are evaluated. 
For DO, a Conformer encoder with two separate transformer decoders is trained jointly on both objectives.
The models are trained for 50 epochs using the AdamW optimizer.
The weight decay is set to 0.01, and the learning rate is set to $10^{-4}$. 
For fine-tuned models, the learning rate is reduced to $10^{-5}$. The batch size is eight, and SpecAugment~\citep{Park_2019} is used for data augmentation. 
Experiments are conducted on a single NVIDIA RTX 3090 GPU. 
The primary reported metric is the character error rate (CER), which is calculated using beam search decoding with a beam size of five.

\begin{table}[h]
\centering
\small
\caption{Model performance reported in CER (\%).}
\label{tab:main_results}
\setlength{\tabcolsep}{2pt} 
\begin{tabular}{lccccc}
\toprule
 & & \multicolumn{2}{c}{\textbf{Korean}} & \multicolumn{2}{c}{\textbf{English}} \\
\cmidrule(lr){3-4} \cmidrule(lr){5-6}
\textbf{Model} & \textbf{Params} & \textbf{Surface} & \textbf{Meaning} & \textbf{Surface} & \textbf{Meaning} \\
\midrule
\multicolumn{6}{l}{\textit{Single-output}} \\
Conformer & 32M & 11.14 & 1.60 & 13.78 & 3.87 \\
Whisper-base & 72M & 10.05 & 4.62 & 11.39 & 0.55 \\
Whisper-small & 244M & 6.76 & 0.54 & 11.20 & 0.27 \\
\midrule
\multicolumn{6}{l}{\textit{Dual-output}} \\
Conformer & 40M & 11.34 & 0.77 & 15.08 & 3.19 \\
\bottomrule
\end{tabular}
\end{table}

\subsection{Results}
Table~\ref{tab:main_results} shows the performance of Korean and English. 
For all models, surface transcription is consistently more difficult than meaning transcription. 
This pattern holds across both languages and all model scales. 
These results align with previous observations that verbatim pronunciation recovery is more challenging than retrieving the standard written form \citep{SARACLAR2004375}. 
Whisper scaling improves both transcription forms; however, the improvement is marginal for English, and the surface-meaning performance gap persists.  

Joint MTL exhibits asymmetric effects across outputs. In both languages, MTL improves meaning transcription but degrades surface transcription. 
The degradation in surface transcription is substantially larger in English than in Korean, while improvements in meaning are comparable. 
This cross-lingual asymmetry under identical training conditions raises a central question addressed through stratified and representation-level analyses.

\begin{figure}[h]
\centering
\includegraphics[width=0.9\linewidth]{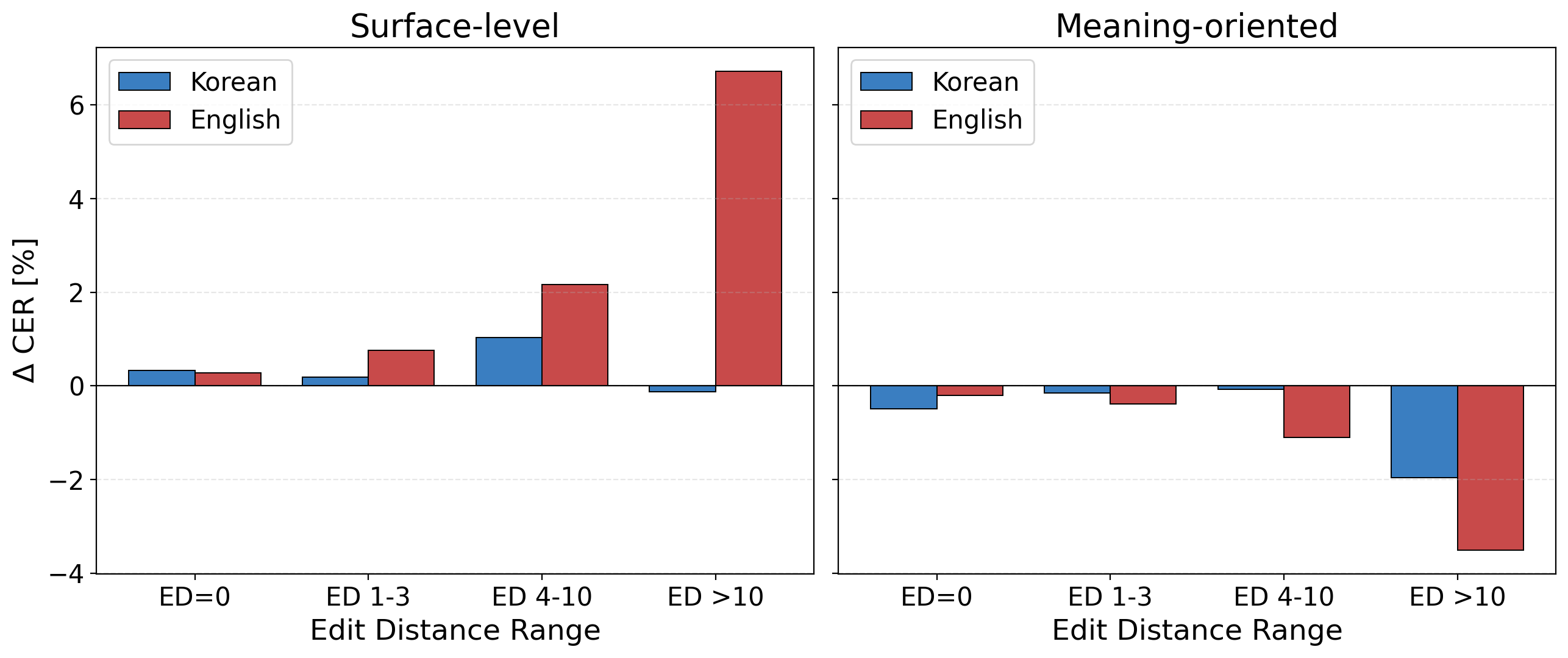}
\caption{CER gap ($\Delta = \text{DO} - \text{SO}$) stratified by ED across languages and outputs. 
Negative values indicate that DO improves over SO, positive values indicate degradation.}
\label{fig:ed_stratified}
\end{figure}

\subsection{Analysis Beyond Distributional Differences}
\label{sec:ed_stratified}

To examine whether cross-lingual asymmetry depends on surface-meaning divergence, the test sets are stratified by ED across two transcriptions. 
For each ED range, the CER gap is compared for surface and meaning outputs separately.  
Figure \ref{fig:ed_stratified} shows the stratified results and reveals two patterns.  

In Korean, MTL effects are minor and inconsistent with divergence. 
The surface gap ranges from +0.19 to +1.03 in the lower ranges and becomes slightly negative at ED$>$10. 
The meaning gap remains small, reaching -1.96 at ED$>$10.  
In English, the CER gap scales in opposite directions with divergence for the two outputs. 
The surface gap increases monotonically from +0.28 at ED$=$0 to +6.72 at ED$>$10.  
The meaning gap strengthens monotonically from -0.20 to -3.51, indicating a systematic surface-meaning trade-off that intensifies with divergence.  
Despite these starkly different MTL behaviors, the underlying ED distributions are similar across languages, as reported in Table \ref{tab:dataset}. 
These findings suggest that the asymmetry cannot be attributed to distributional differences alone, thus motivating a representation-level analysis.

\section{Mechanistic Analysis}
\label{sec:mechanism}
To localize the source of the cross-lingual asymmetry in surface degradation, this section examines encoder representations and decoder representations with cross-output analysis. 
Centered Kernel Alignment (CKA) \citep{kornblith2019similarity} is used to compare the representations of the SO and DO models across layers. 
This process reveals where the representations diverge or converge within the encoder and decoder.

\begin{table}[h]
\caption{Layer-wise CKA between encoder representations.}
\label{tab:encoder_cka}
\centering
\small
\setlength{\tabcolsep}{6.5pt}
\begin{tabular}{c|ccc|ccc}
\toprule
& \multicolumn{3}{c|}{Korean} & \multicolumn{3}{c}{English} \\
& S$_{SO}$ & S$_{SO}$ & M$_{SO}$ & S$_{SO}$ & S$_{SO}$ & M$_{SO}$ \\
Layer & $\updownarrow$ & $\updownarrow$ & $\updownarrow$ & $\updownarrow$ & $\updownarrow$ & $\updownarrow$ \\
& M$_{SO}$ & S$_{DO}$ & M$_{DO}$ & M$_{SO}$ & S$_{DO}$ & M$_{DO}$ \\
\midrule
0  & 0.95 & 0.98 & 0.96 & 0.91 & 0.70 & 0.88 \\
3  & 0.43 & 0.72 & 0.51 & 0.89 & 0.88 & 0.88 \\
6  & 0.56 & 0.74 & 0.60 & 0.75 & 0.72 & 0.84 \\
9  & 0.45 & 0.78 & 0.48 & 0.66 & 0.67 & 0.70 \\
11 & 0.56 & 0.81 & 0.62 & 0.40 & 0.46 & 0.53 \\
\bottomrule
\end{tabular}
\end{table}

\subsection{Encoder Representation}
\label{subsec:encoder}

\begin{table*}[h]
\caption{Layer-wise CKA between decoder representations.}
\label{tab:decoder_cka}
\centering
\small
\setlength{\tabcolsep}{10pt}
\begin{tabular}{c|c|cc|cc|c|cc|cc}
\toprule
& \multicolumn{5}{c|}{\textbf{Korean}} & \multicolumn{5}{c}{\textbf{English}} \\
\cmidrule(lr){2-6} \cmidrule(lr){7-11}
& \textit{Baseline} & \multicolumn{2}{c|}{\textit{Same-output}} & \multicolumn{2}{c|}{\textit{Cross-output}} & \textit{Baseline} & \multicolumn{2}{c|}{\textit{Same-output}} & \multicolumn{2}{c}{\textit{Cross-output}} \\
Layer & S$_{SO}$ & S$_{SO}$ & M$_{SO}$ & S$_{SO}$ & M$_{SO}$ & S$_{SO}$ & S$_{SO}$ & M$_{SO}$ & S$_{SO}$ & M$_{SO}$ \\
& $\updownarrow$ & $\updownarrow$ & $\updownarrow$ & $\updownarrow$ & $\updownarrow$ & $\updownarrow$ & $\updownarrow$ & $\updownarrow$ & $\updownarrow$ & $\updownarrow$ \\
& M$_{SO}$ & S$_{DO}$ & M$_{DO}$ & M$_{DO}$ & S$_{DO}$ & M$_{SO}$ & S$_{DO}$ & M$_{DO}$ & M$_{DO}$ & S$_{DO}$ \\
\midrule
0 & 0.82 & 0.87 & 0.84 & 0.75 & 0.83 & 0.34 & 0.58 & 0.43 & 0.38 & 0.36 \\
3 & 0.52 & 0.75 & 0.85 & 0.52 & 0.56 & 0.52 & 0.72 & 0.59 & 0.49 & 0.54 \\
4 & 0.49 & 0.74 & 0.86 & 0.46 & 0.52 & 0.54 & 0.73 & 0.56 & 0.48 & 0.57 \\
7 & 0.53 & 0.70 & 0.88 & 0.46 & 0.57 & 0.39 & 0.52 & 0.24 & 0.17 & 0.44 \\
\bottomrule
\end{tabular}
\end{table*}

Table~\ref{tab:encoder_cka} reports the CKA between the SO and DO conformer encoders at each layer. 
S and M refer to the surface and meaning outputs, respectively.
S$_{SO}$ refers to the surface SO encoder, M$_{SO}$ refers to the meaning SO encoder, and S$_{DO}$ and M$_{DO}$ refers to the DO encoder viewed through the surface and meaning outputs, respectively.
Notation X $\leftrightarrow$ Y indicates the CKA between two representations X and Y.

In Korean, S$_{SO}$$\leftrightarrow$M$_{SO}$ diverges sharply from Layer 3 onward, indicating that each encoder trains distinct representations.
For the same output, S$_{SO}$$\leftrightarrow$S$_{DO}$ remains consistently high in deeper layers, while M$_{SO}$$\leftrightarrow$M$_{DO}$ shows moderate alignment. 
In English, however, the pattern differs. 
S$_{SO}$$\leftrightarrow$M$_{SO}$ decreases gradually before dropping sharply at the final layer, suggesting that the two SO encoders learn similar representations. 
Only at the final layer does it drop to 0.40, at which point the DO encoder also shows reduced similarity with the two SO encoders.  

Since they produce different outputs, divergence of S$_{SO}$$\leftrightarrow$M$_{SO}$ at the final layer was expected. 
However, pairs that target the same output also exhibit a similar drop in similarity, except for Korean S$_{SO}$$\leftrightarrow$S$_{DO}$. 
This suggests that, despite targeting the same output as the SO counterpart, the DO encoder fails to develop distinct representations. 
We refer to this phenomenon as "encoder-level entanglement." 
It is inconsistent with the meaning improvement observed in Table~\ref{tab:main_results}, which motivates analyzing how these representations are processed at the decoder level.

\subsection{Decoder Representation}
\label{subsec:decoder}

Table~\ref{tab:decoder_cka} shows the CKA between the SO and DO decoders, which are organized into three groups. 
The S$_{SO}$$\leftrightarrow$M$_{SO}$ comparison illustrates how the two SO decoders compare.
Same-output comparisons (S$_{SO}$$\leftrightarrow$S$_{DO}$ and M$_{SO}$$\leftrightarrow$M$_{DO}$) compare each DO decoder with the SO decoder targeting the same output.
The cross-output comparisons (S$_{SO}$$\leftrightarrow$M$_{DO}$ and M$_{SO}$$\leftrightarrow$S$_{DO}$) instead pair each DO decoder with the opposing SO decoder.

In Korean, the baseline S$_{SO}$$\leftrightarrow$M$_{SO}$ diverges from 0.82 in the input layer to 0.53 in the final layer. 
This shows that the two SO decoders develop distinct representations. 
Same-output S$_{SO}$$\leftrightarrow$S$_{DO}$ decreases gradually with depth, while the M$_{SO}$$\leftrightarrow$M$_{DO}$ approaches 0.88 at deeper layers. 
Same-output values consistently exceed cross-output values across all layers, indicating that Korean DO decoders align with their respective outputs.

However, the pattern is different in English. 
The baseline S$_{SO}$$\leftrightarrow$M$_{SO}$ remains low from the beginning, peaking only in the middle layers before dropping in the final layer. 
The same-output S$_{SO}$$\leftrightarrow$S$_{DO}$ increases in the middle layers, then falls to 0.52 at Layer 7. 
Meanwhile, M$_{SO}$$\leftrightarrow$M$_{DO}$ drops to 0.24. 
Cross-output S$_{SO}$$\leftrightarrow$M$_{DO}$ drops even further, reaching 0.17. 
These results confirm that M$_{DO}$ constructs a representation distinct from all other decoders, effectively bypassing the entangled encoder.
S$_{DO}$ lacks this flexibility because it must remain connected to the encoder to produce frame-aligned transcription. 
This asymmetric flexibility explains why English exhibits significant surface degradation alongside moderate meaning improvement under MTL.

Overall, Korean decoders develop representations specific to the output, consistent with the encoder-level separation. 
The meaning decoder strengthens alignment beyond what the encoder achieves. 
However, the entangled English encoder provides no such foundation, limiting decoder-level adaptation. 
These findings suggest that encoder-level separation is a prerequisite for effective decoder-level specialization.

\section{Conclusion}
This paper investigated joint MTL for DO L2 ASR across Korean and English. 
MTL improves meaning but degrades surface transcription. 
The degree of degradation is substantially larger in English, as surface-meaning divergence increases. 
CKA analysis reveals that the Korean encoder learns disentangled representations, whereas the English encoder exhibits entanglement. 
At the decoder level, Korean decoders build on this separation, further strengthening output-specific alignment. 
However, in English, the meaning decoder compensates by constructing a distinct representation that bypasses the entangled encoder, while the surface decoder cannot. 
These findings suggest that encoder-level separation is a prerequisite for effective decoder-level specialization, motivating MTL approaches that mitigate encoder-level entanglement in DO L2 ASR. 
Promising directions include sparse decomposition, adversarial training, and gating mechanisms, along with validation through complementary similarity metrics beyond CKA.

\section*{Acknowledgements}
This work was supported by the National Research Foundation of Korea (NRF) grant funded by the Korea government (MSIT) (RS-2026-25492127).

\bibliography{L2L}
\bibliographystyle{icml2026}

\newpage
\appendix
\onecolumn


\end{document}